\documentclass{llncs}

\usepackage[utf8]{inputenc}
\usepackage[T1]{fontenc}

\usepackage[final]{graphicx}
\usepackage{epstopdf}
\usepackage[labelsep=period]{caption}
\usepackage[hyphens]{url}
\usepackage[russian,english]{babel}

\usepackage{amsmath}
\usepackage{adjustbox}
\usepackage{tikz}
\usetikzlibrary{shapes.geometric, arrows}
\tikzstyle{block} = [rectangle, rounded corners, minimum width=3cm, minimum height=1cm,text centered, draw=black]
\tikzstyle{arrow} = [thick,->,>=stealth]

\title{Impulsive noise removal from color images with morphological filtering}

\author{
Alexey Ruchay
\and
Vitaly Kober
}

\institute{
Department of Mathematics, Chelyabinsk State University, Russia\\
\email{ran@csu.ru}\\
}

\DeclareMathOperator*{\argmin}{argmin}
\graphicspath{{./New/}}

\begin{document}
\maketitle

%%%%%%%%%%%%%%%%%%%%%%%%%%%%%%%%%%%%%%%%%%%%%%%%%%%%%%%%%%%%%
\begin{abstract}
This paper deals with impulse noise removal from color images. The proposed noise removal algorithm employs a novel approach with morphological filtering for color image denoising; that is, detection of corrupted pixels and removal of the detected noise by means of morphological filtering. With the help of computer simulation we show that the proposed algorithm can effectively remove impulse noise. The performance of the proposed algorithm is compared in terms of image restoration metrics and processing speed with that of common successful algorithms.
\end{abstract}

\keywords{color image, impulsive noise removal, denoising, morphological filtering.}

%%%%%%%%%%%%%%%%%%%%%%%%%%%%%%%%%%%%%%%%%%%%%%%%%%%%%%%%%%%%%
\section{INTRODUCTION}
Color image processing has received much attention in the last years \cite{Kober06}. Digital image processing algorithms are generally sensitive to noise. A color image is treated as a mapping $Z_2 \to Z_3$ that assigns to a point $x=(i,j)$ on the image plane a three-dimensional vector $(x^r,x^g,x^b)$, where the superscripts correspond to the red, green, and blue color image channels. In this way, a color image is considered as a two-dimensional vector field, and each vector has three color components.

\begin{figure}[!tb]
\begin{center}
%\begin{tabular}{c}
%\includegraphics[height=3cm]{shema2}
%\end{tabular}
\resizebox{!}{3cm}{
\begin{tikzpicture}[node distance=2cm]
\node [block] (start){Corrupted image};
\node [block, right of=start, xshift=2cm] (item1){Noise detection};
\node [block, above right of=item1, xshift=3cm](item2) {Noise filtering};
\node [block, below right of=item1, xshift=3cm](item3) {No filtering};
\node [block, below right of=item2, xshift=3cm](end) {Restored image};
\draw [arrow] (start) -- (item1);
\draw [arrow] (item1) -- (item2);
\draw [arrow] (item1) -- (item3);
\draw [arrow] (item2) -- (end);
\draw [arrow] (item3) -- (end);
\end{tikzpicture}
}
\end{center}
\caption[example]
{\label{fig:RuchSchema} Switching filtering scheme.}
\end{figure}
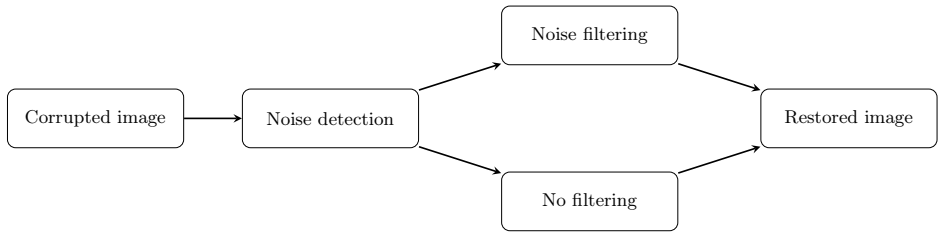

The most popular algorithms for removal of impulsive noise in color images utilize the ordering of pixels belonging to a local window $W$ \cite{Dinet07}, and assign a dissimilarity measure to each color pixel from the window. Several switching techniques are proposed \cite{Smolka15,SmolkaJ06} to adapt parameters of filters to the processed image. A switching algorithm verifies the following hypothesis: is the central pixel of window $W$ affected by noise? If the central pixel is corrupted by noise then it is replaced by the output of a local robust filter; otherwise, it is left unchanged (see Fig. \ref{fig:RuchSchema}). One of efficient switching schemes is referred to as the sigma vector median filter (SVMF) \cite{SmolkaJ06}.

The performance of a switching filtering depends mainly on the impulse noise detection. If the detector fails to identify corrupted pixels, the performance of the algorithm yields errors of missed impulse noise. On the other hand, if the detector wrongly identifies uncorrupted pixels as noisy, the performance of the algorithm yields false impulse noise errors. In the both cases the overall performance of image restoration is poor. The performance of switching filtering algorithms can be compared with various image restoration measures \cite{Smolka15,SmolkaJ06}. In this paper, type I and type II errors are used to characterize the performance of tested algorithms. A type I error occurs when the algorithm asserts something that is absent, a false hit. A type I error is called false positive (FP). A type II error occurs when the algorithm fails to assert what is present, a miss. A type II error is called false negative (FN).

Mathematical morphology describes the shape and structure of certain objects, and is used to extract the useful components in the image. It is utilized for image filtering, image segmentation, image measurement, area filling and so on \cite{Soille,najman}.
In the image denoising aspect, we can get fairly good effect by applying the gray morphology, having the characteristics of nonlinearity and parallelism \cite{Jakhar,Kimori}.

In this paper a novel approach to color image denoising by morphological filtering is proposed. With the help of computer simulation we show that the proposed algorithm can effectively remove impulse noise. The performance of the proposed algorithm is compared in terms of image restoration metrics with that of common successful algorithms \cite{Ruchay}.

The paper is organized as follows. In Section 2, we describe the proposed switching algorithm by morphological filtering. Section 3 describes impulsive noise models. Computer simulation results are provided in Section 4. Finally, Section 5 summarizes our conclusions.

%%%%%%%%%%%%%%%%%%%%%%%%%%%%%%%%%%%%%%%%%%%%%%%%%%%%%%%%%%%%%
\section{Proposed algorithm}

A common impulse noise removal algorithm is based on the reduced vector ordering, which assigns a dissimilarity measure to each color pixel $x_i$ from the local window $W=\{x_1,x_2,\ldots,x_n\}$ of the size $n=9$. Let $\rho(x_i,x_j)$ be the distance between two vectors $x_i$, $x_j$, then the inner product is defined as
\begin{equation}
\label{eq:Ruch1}
d_i = \sum_{j=1}^{n}\rho(x_i,x_j), x_j \in W.
\end{equation}
The meaning of the product is the distance associated with the central pixel $x_i$ inside the filtering window $W$. The ordering of the distances as $d_{(1)} \leq d_{(2)} \leq \ldots \leq d_{(n)},$ implies the same ordering to the corresponding vectors $x_j$
$x_{(1)} \leq x_{(2)} \leq \ldots \leq x_{(n)}.$

The original value of the central pixel $x_1$ in the window $W$ is being replaced by $x_{(1)}$ which means that
\begin{equation}
\label{eq:Ruch2}
x_{(1)} = \argmin\limits_{x_i \in W} d_i.
\end{equation}

This concept of replacing is a common way to define the mean scale in vector spaces. It is called the Vector Median Filter (VMF) \cite{Khryashchev}. Most commonly the $L_2$ metric is used for the design of the VMF
$$\rho(x_i,x_j)=\left(\sum_{k=1}^3 (x^k_i - x^k_j)^2\right)^{1/2}.$$

\subsection{Rank weighted vector median filter}

The reduced ordering schemes are based on the sum of the dissimilarity measures between a given pixel and all other pixels from the filtering window $W$ \cite{Smolka15}. In this way, the output of the VMF is the pixel whose average distance to other pixels is minimized.

The distances $d_{ij}=\|x_i - x_j \|$ between the pixel $x_i$ and all other pixels $x_j$ belonging to $W, (j=1,\ldots,n)$ can be ordered as
$d_{i1},d_{i2}, \ldots, d_{in} \rightarrow d_{i(1)} \leq d_{i(2)} \leq \cdots \leq d_{i(n)}$,
and the ranks of the ordered distances can be used for building the cumulated distances in Eq.~(\ref{eq:Ruch1}).

Let $r$ denote the rank of a given distance, and $d_{i(r)}$ stand for the corresponding distance value. So, instead of the aggregated distances in Eq.~(\ref{eq:Ruch1}) we can build a weighted sum of distances, utilizing the distance ranks as
$$\Delta_i=\sum_{i=1}^n f(r) d_{i(r)},$$
where $f(r)$ is a decreasing weighting function of the distance rank $r$, like $f(r) = 1$, $f(r) = 1/r$ and $f(r) = 1/r^2$. The $f(r) = 1/r$ weights for the design of the adaptive switching filter is recommended in \cite{Smolka15}.

Then, the rank weighted sum of distances calculated for each pixel belonging to $W$ can be sorted and a new sequence of vectors can be obtained
$\Delta_{1},\Delta_{2}, \ldots, \Delta_{n} \rightarrow x^*_{(1)} \leq x^*_{(2)} \leq \cdots \leq x^*_{(n)}$,
where the vector $x^*_{(1)}$ is the output of the rank weighted vector median filter (RWVMF) \cite{Smolka15}.

Similarly to Eq.~(\ref{eq:Ruch2}) the RWVMF output $x^*_{(1)}$ an be defined as
$$x^*_{(1)} = \argmin\limits_{x_i \in W} \sum_{r=1}^n f(r) d_{i(r)}.$$

The structure of the switching filter is defined \cite{Smolka15} as follows. If the difference $\Delta_1-\Delta_{(1)}$ exceeds a threshold value $\alpha$, then a pixel is declared as corrupted by a impulsive noise; otherwise, it is treated as uncorrupted
\begin{equation}
\label{eq:Ruch3}
y_1 =
 \begin{cases}
   x_{AMF}, &\text{if} \quad \Delta_1-\Delta_{(1)}>\alpha,\\
   x_1, &\text{otherwise},\\
 \end{cases}
\end{equation}
where $y_1$ is the switching filter output, $x_1$ is the central pixel of the filtering window $W$ and $x_{AMF}$ is the Arithmetic Mean Filter (AMF) output computed over the pixels declared by the detector as uncorrupted. Extensive experiments revealed that very good denoising results can be achieved using the following switching filter:
$$
y_1 =
 \begin{cases}
   x_{VMF}, &\text{if} \quad \Delta_1-\Delta_{(1)}>\alpha,\\
   x_1, &\text{otherwise},\\
 \end{cases}
$$
where $x_{VMF}$ is the standard VMF output computed for all the pixels in the filtering window $W$.

Detection noise method \verb"DetectionMethod1" for pixel $x_1$ of the filtering window $W$ in RWVMF in Eq.~(\ref{eq:Ruch3}) can be defined as follows:
\begin{equation}
\label{eq:Ruch4}
\verb"DetectionMethod1"(x_1) =
 \begin{cases}
   1, &\text{if} \quad \Delta_1-\Delta_{(1)}>\alpha,\\
   0, &\text{otherwise},\\
 \end{cases}
\end{equation}
where 1 means the successful detection of noise and 0 means no noise detected.

We propose the following modification of the noise detection given in Eq.~(\ref{eq:Ruch4}). This detection noise method \verb"DetectionMethod2" for pixel $x_1$ of the filtering window $W$ in RWVMF can be defined as follows:
$$
\verb"DetectionMethod2"(x_1) =
 \begin{cases}
   1, &\text{if} \quad \Delta_{(1)}>\alpha,\\
   0, &\text{otherwise}.\\
 \end{cases}
$$

The detection method \verb"DetectionMethod1" and \verb"DetectionMethod2" use the predefined parameter $\alpha$.

\subsection{Fast peer group filter}
Recently, a peer group filter has been proposed \cite{SmolkaB05,Smolka16}. The peer group associated with the central pixel of the window denotes a set of such pixels whose distance to the central pixel does not exceed a predefined threshold. The Fast Peer Group Filter (FPGF) replaces the center of the filtering window with the VMF output when a specified number of the smallest distances between the central pixel and its neighbors differ not more than a predefined threshold.

Let vector components $x_i \in [0,1]$ represent the color channel values in a given color space quantified into the integer domain. In the first step, the size of the peer group, or in other words, the number of close neighbors of the central pixel of the filtering window $x_1$ is determined. A pixel $x_i \neq x_1$ belonging to $W$ is a close neighbor of $x_1$, if the normalized Euclidean distance $d(x_i,x_1)$ in a given color space is less than a predefined threshold valued $d\in[0,1]$.

In the RGB color space, the peer group size denoted as $m_k$ is the number of pixels from $W$ contained in a sphere with radius $d$ centered at pixel $x_k$
$m_k=\#\{x_j \in W:  \|x_j-x_k\|< d \},$
where $\#$ denotes the cardinality and $\|\cdot\|$ stands for the Euclidean norm.

If the peer group size of the central pixel $x_1$ of the filtering window $W$ is $m_1 \leq 2$, then this pixel is treated as an outlier. The structure of the switching filter can be defined as follows:
\begin{equation}
\label{eq:Ruch5}
y_1 =
 \begin{cases}
   x_{VMF}, &\text{if} \quad m_k \leq k,\\
   x_1, &\text{otherwise},\\
 \end{cases}
\end{equation}
where $x_{VMF}$ is standard VMF output computed for all the pixels of window $W$, and $k$ is a parameter that determines the minimal size of the peer group.

Detection noise method \verb"DetectionMethod3" for the pixel $x_1$ of the window $W$ in Eq.~(\ref{eq:Ruch5}) can be defined as follows:
\begin{equation}
\label{eq:Ruch6}
\verb"DetectionMethod3"(x_1) =
 \begin{cases}
   1, &\text{if} \quad m_k \leq k,\\
   0, &\text{otherwise}.\\
 \end{cases}
\end{equation}
The detection method \verb"DetectionMethod3" has parameter $d$ and $k$.

We propose a modification of the detection noise method \verb"DetectionMethod3" in Eq.~(\ref{eq:Ruch6}). The proposed method \verb"DetectionMethod4" utilizes iteratively the detection noise method \verb"DetectionMethod3". At first step \verb"DetectionMethod3" is used with parameters $d=0.25$ and $k=3$. This step corresponds to a preliminary detection of noise. Then, the \verb"DetectionMethod3" is iteratively used with modified parameters $d$ and $k$. Experiments showed that good denoising results can be achieved using the proposed detection method.

\subsection{Morphological filter}
Morphological processing is constructed with operations on sets of pixels. Binary morphology uses only set membership and is indifferent to the value, such as gray level or color, of a pixel. We will deal here only with morphological operations for binary images. Therefore we use a threshold operation \verb"BW(A, level)" to convert the grayscale image $A$ to a binary image. The output image replaces all pixels in the input image with than the threshold \verb"level" by 1 (white) and replaces all other pixels with 0 (black).

The operation intersection $A \cap B$ produces a set that contains the elements in both $A$ and $B$.
The operation union $A \cup B$ produces a set that contains the elements of both $A$ and $B$.
The complement $A^c$ is the set of elements that are not contained in $A$.
The difference of two sets $A$ and $B$, denoted by $A - B$ is $A \cap B^c$.
A standard morphological operation is the reflection of all of the points in a set $\widehat{A}$ about the origin of the set $A$.

Dilation and erosion are basic morphological processing operations.
Let $A$ be a set of pixels and let $B$  be a structuring element. Let $(\widehat{B})_s$ be the reflection of $B$ about its origin and followed by a shift by $s$. Dilation operation is the set of all shifts that satisfy the following: $A \oplus B = \{s | ((\widehat{B})_s \cap A) \subseteq A \}$.
Erosion operation is the set of all shifts that satisfy the following: $A \ominus B = \{ s | (B)_s \subseteq A \}$.

Closing operation is a dilation followed by an erosion: $A \circ B = (A \oplus B) \ominus B$.
Opening operation is an erosion followed by a dilation: $A \bullet B =(A \ominus B) \oplus B$.

Morphological "bottom hat" operation is an image minus the morphological closing of an image: $A \star B = ((A \oplus B) \ominus B) - A$.

Morphological "remove" operation is a removing interior pixels of an image $A$, written as $\verb"remove"(A)$. This operation sets a pixel to 0 if all its 4-connected neighbors are 1, thus leaving only the boundary pixels on.

Let a color image $X$ be three-dimensional vector $(x^r, x^g, x^b)$ each channel is processed individually. We propose the following  \verb"DetectionMethod5"(X) method of the noise detection for color image $X$ with a morphological filter. The output of this method is $$M=M_1 \cup M_2 \cup M_3 \cup M_4 \cup M_5.$$

$$M_1=\bigcup_{i=r,g,b}(\verb"set1"(x^i,mset) \star B) \bigcup_{i=r,g,b}(\verb"set2"(x^i,pset) \star B),$$
$$M_2=\verb"remove"\left(\bigcup_{i=r,g,b}\verb"BW"(\verb"set2"(x^i,pset),\verb"level")\right),$$
$$M_3=\verb"remove"\left(\bigcup_{i=r,g,b}\verb"BW"(\verb"set3"(x^i,mset),\verb"level")\right),$$
$$M_4 =\left( \verb"BW"(\verb"rgb2gray"(X),level) \star B \right),$$
$$M_5 = \left( \verb"BW"(\verb"rgb2gray"(\verb"set2"(X,\verb"pset")),level) \star B \right),$$
where $B$ is the standard structuring element, \verb"set1(A,mset)" is subtraction of all the pixels \verb"A" value \verb"mset", \verb"set2(A,pset)" is subtraction of the values \verb"pset" of all the pixels \verb"A", \verb"set3(A,mset)" is addition of all the pixels \verb"A" value \verb"mset", \verb"rgb2gray" is conversion of the color image $X$ to the grayscale intensity image.

The detection method \verb"DetectionMethod5" uses the parameters: \verb"pset", \verb"mset", \verb"level".

%%%%%%%%%%%%%%%%%%%%%%%%%%%%%%%%%%%%%%%%%%%%%%%%%%%%%%%%%%%%%
\section{Model of impulse noise}

Color images may be contaminated by various types of impulse noise \cite{Khryashchev,Singh,Venkatesan,Smolka15}. Impulse noise corruption often occurs in digital image acquisition or transmission process as a result of photo-electronic sensor faults or channel bit errors. Image transmission noise may be caused by various sources, such as car ignition systems, industrial machines in the vicinity of the receiver, switching transients in power lines, lightning in the atmosphere and various unprotected switches. This type of transmission noise is often modeled as impulse noise. Let us consider models of impulse noise used for computer simulation. Let  $X_i$  be the vector characterizing a pixel of a noisy image, $q$ be the vector describing one of the noise models, $x_i$ be the noise-free color vector, $p$ be the probability of impulse noise occurrence. Each tested image can be corrupted with different probabilities, that is, $p \in \{0.1,0.2,0.3\}$. Depending on the type of vector $q$, either fixed-valued or random-valued impulse noise models are considered.

Assume that channels are corrupted independently (CI). So, we use the following models of impulse noise:
$$
X_i =
 \begin{cases}
   (q_1,x_i^g,x_i^b), &\text{with probability} \quad p(1-p)^2,\\
   (x_i^r,q_2,x_i^b), &\text{with probability} \quad p(1-p)^2,\\
   (x_i^r,x_i^g,q_3), &\text{with probability} \quad p(1-p)^2,\\
   (q_1,q_2,x_i^b),   &\text{with probability} \quad p^2(1-p),\\
   (x_i^r,q_2,q_3), &\text{with probability} \quad p^2(1-p),\\
   (q_1,x_i^g,q_3), &\text{with probability} \quad p^2(1-p),\\
   (q_1,q_2,q_3), &\text{with probability} \quad p^3,\\
   (x_i^r,x_i^g,x_i^b), &\text{with probability} \quad (1-p)^3,\\
 \end{cases}
$$
where $q_1,q_2,q_3$ are spatially uniform distributed independent random variables with the probability of $p$. The  corrupted pixels can be defined in different manner; that is, CI1 means that they take values of either 0 or 255; CI2 means that corrupted pixel is a random variable with uniform distribution in the interval of $[0, 255]$; CI3 means that corrupted pixel is a random variable with uniform distribution in the intervals of $[0, 55]$ and $[200, 255]$. Additionally, we introduce a model CT when all channels of the color image are contaminated simultaneously by impulsive noise as follows:
$$
X_i =
 \begin{cases}
   (q_1,q_2,q_3), &\text{with probability} \quad p,\\
   (x_i^r,x_i^g,x_i^b), &\text{with probability} \quad (1-p).\\
 \end{cases}
$$
The corrupted pixels can be defined in different manner as CT1, CT2, CT3.

%%%%%%%%%%%%%%%%%%%%%%%%%%%%%%%%%%%%%%%%%%%%%%%%%%%%%%%%%%%%%
\section{Computer Simulation}
The performance of the detection methods \verb"DetectionMethod(1-5)" is compared with respect to FP and FN errors. Since FP and FN errors depend on the parameters of the detection methods, then the receiver operating characteristic (ROC) curve as a function of FP and FN errors is utilized. The parameters of detection methods can be chosen from the ROC curve to provide the minimum FP and FN errors.
%In Fig. \ref{fig:RuchFig31}, \ref{fig:RuchFig32}, \ref{fig:RuchFig33} the ROC curves for detection methods \verb"DetectionMethod(1-5)" with the type of noise CI1-3, CT1-3, $p=0.3$ are shown.
Minimum FP and FN errors for all tested methods \verb"DetectionMethod(1-5)" with the type of noise CI1-3, CT1-3, $p=0.1,0.2,0.3$ are summarized in Table \ref{tab:RuchTab1}.
One can be observe that the proposed method \verb"DetectionMethod5" detects noise very well comparing with other detection techniques.
The algorithm of the removal of impulsive noise by a switching filter can be defined as follows:
$$y_1 =
 \begin{cases}
   x_{VMF}, &\text{if} \quad \verb"DetectionMethod"(x_1)=1,\\
   x_1, &\text{otherwise},\\
 \end{cases}
$$
where $x_{VMF}$ is the standard VMF output computed for all the pixels of the window $W$.

\begin{table}[!tb]
\caption{Minimum FP and FN errors for DetectionMethod(DM) 1-5 with type of noise (TN) CI1-3, CT1-3, $p=0.1,0.2,0.3$.}
\label{tab:RuchTab1}
\centering
\begin{adjustbox}{width=\textwidth}
\begin{tabular}{|l|c|c|c|c|c|c|c|c|c|c|}
\hline
\rule[-1ex]{0pt}{3.5ex} TN & FPDM1 & FNDM1 & FPDM2 & FNDM2 & FPDM3 & FNDM3 & FPDM4 & FNDM4 & FPDM5 & FNDM5\\
\hline
\rule[-1ex]{0pt}{3.5ex}  CI1 0.1 & 0.037  & 0.094  & 0.094  & 0.062  & 0.067  & 0.099  & 0.021  & 0.075  & \textbf{0.033}  & \textbf{0}      \\
\hline
\rule[-1ex]{0pt}{3.5ex}  CI2 0.1 & 0.100  & 0.145  & 0.099  & 0.107  & 0.115  & 0.212  & \textbf{0.063}  & \textbf{0.123}  & 0.113  & 0.263  \\
\hline
\rule[-1ex]{0pt}{3.5ex}  CI3 0.1 & 0.090  & 0.107  & 0.097  & 0.087  & 0.121  & 0.155  & 0.048  & 0.092  & \textbf{0.075}  & \textbf{0.034}  \\
\hline
\rule[-1ex]{0pt}{3.5ex}  CT1 0.1 & 0.016  & 0.013  & 0.085  & 0.017  & 0.043  & 0.007  & 0.004  & 0.009  & \textbf{0.03}   & \textbf{0}      \\
\hline
\rule[-1ex]{0pt}{3.5ex}  CT2 0.1 & 0.042  & 0.022  & 0.086  & 0.036  & 0.089  & 0.062  & 0.027  & 0.043  & \textbf{0.111}  & \textbf{0.001}  \\
\hline
\rule[-1ex]{0pt}{3.5ex}  CT3 0.1 & 0.049  & 0.043  & 0.086  & 0.042  & 0.078  & 0.064  & 0.027  & 0.044  & \textbf{0.062}  & \textbf{0.004}  \\
\hline
\rule[-1ex]{0pt}{3.5ex}  CI1 0.2 & 0.143  & 0.098  & 0.104  & 0.109  & 0.312  & 0.065  & 0.05   & 0.088  & \textbf{0.037}  & \textbf{0}      \\
\hline
\rule[-1ex]{0pt}{3.5ex}  CI2 0.2 & 0.175  & 0.134  & 0.118  & 0.144  & 0.181  & 0.207  & \textbf{0.083}  & \textbf{0.112}  & 0.205  & 0.218  \\
\hline
\rule[-1ex]{0pt}{3.5ex}  CI3 0.2 & 0.173  & 0.101  & 0.108  & 0.125  & 0.184  & 0.136  & 0.064  & 0.102  & \textbf{0.092}  & \textbf{0.028}  \\
\hline
\rule[-1ex]{0pt}{3.5ex}  CT1 0.2 & 0.022  & 0.086  & 0.060  & 0.083  & 0.152  & 0.044  & 0.007  & 0.071  & \textbf{0.131}  & \textbf{0}      \\
\hline
\rule[-1ex]{0pt}{3.5ex}  CT2 0.2 & 0.023  & 0.047  & 0.062  & 0.096  & 0.054  & 0.052  & 0.010  & 0.034  & \textbf{0.158}  & \textbf{0.026}  \\
\hline
\rule[-1ex]{0pt}{3.5ex}  CT3 0.2 & 0.031  & 0.110  & 0.062  & 0.099  & 0.078  & 0.110  & 0.023  & 0.083  & \textbf{0.068}  & \textbf{0.004}  \\
\hline
\rule[-1ex]{0pt}{3.5ex}  CI1 0.3 & 0.397  & 0.087  & 0.159  & 0.132  & 0.659  & 0.036  & 0.167  & 0.107  & \textbf{0.043}  & \textbf{0}      \\
\hline
\rule[-1ex]{0pt}{3.5ex}  CI2 0.3 & 0.289  & 0.237  & 0.15   & 0.304  & 0.280  & 0.311  & \textbf{0.161}  & \textbf{0.261}  & 0.14   & 0.446  \\
\hline
\rule[-1ex]{0pt}{3.5ex}  CI3 0.3 & 0.294  & 0.153  & 0.067  & 0.240  & 0.438  & 0.079  & 0.163  & 0.122  & \textbf{0.158}  & \textbf{0.024}  \\
\hline
\rule[-1ex]{0pt}{3.5ex}  CT1 0.3 & 0.041  & 0.112  & 0.098  & 0.084  & 0.342  & 0.009  & 0.020  & 0.065  & \textbf{0.028}  & \textbf{0}      \\
\hline
\rule[-1ex]{0pt}{3.5ex}  CT2 0.3 & 0.041  & 0.113  & 0.103  & 0.102  & 0.169  & 0.030  & 0.024  & 0.075  & \textbf{0.082}  & \textbf{0.023}  \\
\hline
\rule[-1ex]{0pt}{3.5ex}  CT3 0.3 & 0.048  & 0.210  & 0.106  & 0.111  & 0.197  & 0.138  & 0.033  & 0.172  & \textbf{0.087}  & \textbf{0.005}  \\
\hline
\end{tabular}
\end{adjustbox}
\end{table}

We use the mean square error (MSE) and the peak signal to noise ratio (PSNR) as measures of restoration quality. They are defined as
$$MSE = \frac{1}{3N} \sum_{i=1}^N \sum_{k=1}^3 (x_i^k - y_i^k)^2, \quad PSNR = 20 \log_{10} \left( \frac{255}{\sqrt{MSE}}\right),$$
where $x_i^k$, $k=1,2,3$ are the component of the original image, and $y_i^k$ are the restored components.

In order to provide comparison of noise removal techniques taking into account subjective human evaluation, we use FSIMc \cite{Zhang11}, SR-SIM \cite{Zhang12} and IFS \cite{Chang15} quality metrics which are suitable for inspection of color images. The results of impulsive noise removal presented in Tables \ref{tab:RuchTab2} and \ref{tab:RuchTab3} show that proposed method \verb"DetectionMethod5" with morphological filtering achieves the best performance with respect to the all considered quality color image metrics.

\begin{table}[!tb]
\caption{ Comparison of efficiency of denoising by switching filter based on DetectionMethod(DM) 1-5 using quality measure (QM) PSNR, MSE, IFS, FSIM, SRSIM  with type of noise CI1-3, CT1-3, $p=0.1,0.2$.}
\label{tab:RuchTab2}
\begin{center}
\begin{adjustbox}{width=\textwidth}
\begin{tabular}{|l|c|c|c|c|c|c|c|c|c|}
\hline
\rule[-1ex]{0pt}{3.5ex} QM & CI1 0.1 & CI2 0.1 & CI3 0.1 & CT1 0.1 & CT2 0.1 & CT3 0.1 & CI1 0.2 & CI2 0.2 & CI3 0.2 \\
\hline
\rule[-1ex]{0pt}{3.5ex} PSNR DM1 & 28.986 & 27.685  & 27.832  & 30.692  & 30.259 & 29.145 & 24.262  & 24.732  & 24.470  \\
\hline
\rule[-1ex]{0pt}{3.5ex} PSNR DM2 & 28.529 & 30.154  & 29.715  & 30.345  & 31.808 & 30.947 & 23.318  & 25.531  & 24.721  \\
\hline
\rule[-1ex]{0pt}{3.5ex} PSNR DM3 & 28.294 & 27.770  & 27.843  & 30.411  & 28.881 & 28.822 & 23.395  & 24.559  & 24.734  \\
\hline
\rule[-1ex]{0pt}{3.5ex} PSNR DM4 & 29.852     & \bf 30.570  & 29.983  & 32.408   & \bf 31.853 & 31.600 & 25.687  & \bf 25.620  & 25.689  \\
\hline
\rule[-1ex]{0pt}{3.5ex} PSNR DM5 & \bf 32.025 & 27.754  & \bf 30.835  & \bf 34.254  & 31.052 & \bf 31.664 & \bf 27.482  & 24.784  & \bf 27.042  \\
\hline
\rule[-1ex]{0pt}{3.5ex} MSE DM1 & 82.109 & 110.78  & 107.10  & 55.447  & 61.252 & 79.159 & 243.70  & 218.69  & 232.31  \\
\hline
\rule[-1ex]{0pt}{3.5ex} MSE DM2 & 91.234 & 62.748  & 69.433  & 60.050  & 42.878 & 52.275 & 302.83  & 181.94  & 219.24  \\
\hline
\rule[-1ex]{0pt}{3.5ex} MSE DM3 & 96.306 & 108.65  & 106.85  & 59.141  & 84.117 & 85.278 & 297.55  & 227.59  & 218.58  \\
\hline
\rule[-1ex]{0pt}{3.5ex} MSE DM4 & 67.270 & \bf 60.363  & 66.304  & 37.343  & \bf 42.723 & 51.624 & 175.53  & \bf 178.23  & 175.44 \\
\hline
\rule[-1ex]{0pt}{3.5ex} MSE DM5 & \bf 40.798 & 109.067 & \bf 53.659  & \bf 24.418  & 51.041 & \bf 44.328 & \bf 116.128 & 216.119 & \bf 128.499 \\
\hline
\rule[-1ex]{0pt}{3.5ex} IFS DM1 & 0.9636  & 0.9493   & 0.9532   & 0.9796  & 0.9717  & 0.9578  & 0.9005   & 0.9025   & 0.9029   \\
\hline
\rule[-1ex]{0pt}{3.5ex} IFS DM2 & 0.9576  & 0.9675   & 0.9668   & 0.9780  & 0.9754  & 0.9643  & 0.8932   & 0.9194   & 0.9155   \\
\hline
\rule[-1ex]{0pt}{3.5ex} IFS DM3 & 0.9546  & 0.9498   & 0.9509   & 0.9754  & 0.9568  & 0.9501  & 0.8918   & 0.9040   & 0.9113   \\
\hline
\rule[-1ex]{0pt}{3.5ex} IFS DM4 & 0.9690 & \bf 0.9687 & 0.9685 & 0.9860 & \bf 0.9887 & 0.9677 & 0.9256 & \bf 0.9199 & 0.9228 \\
\hline
\rule[-1ex]{0pt}{3.5ex} IFS DM5 & \bf 0.9756 & 0.9514 & \bf 0.9696  & \bf 0.9876  & 0.9656  & \bf 0.9699  & \bf 0.9437   & 0.9141   & \bf 0.9337   \\
\hline
\rule[-1ex]{0pt}{3.5ex} FSIM DM1 & 0.9789  & 0.9669   & 0.9680   & 0.9865  & 0.9834  & 0.9730  & 0.9378   & 0.9349   & 0.9380   \\
\hline
\rule[-1ex]{0pt}{3.5ex} FSIM DM2 & 0.9772  & 0.9767   & 0.9749   & 0.9856  & 0.9860  & 0.9775  & 0.9330   & 0.9422   & 0.9435   \\
\hline
\rule[-1ex]{0pt}{3.5ex} FSIM DM3 & 0.9744  & 0.9669   & 0.9654   & 0.9856  & 0.9753  & 0.9681  & 0.9316   & 0.9302   & 0.9424   \\
\hline
\rule[-1ex]{0pt}{3.5ex} FSIM DM4 & 0.9823 & \bf 0.9769 & 0.9753 & 0.9912 & \bf 0.9864 & 0.9783 & 0.9542 & \bf 0.9436 & 0.9520\\
\hline
\rule[-1ex]{0pt}{3.5ex} FSIM DM5 & \bf 0.9845  & 0.9694 & \bf 0.9769  & \bf 0.9928  & 0.977 & \bf 0.9785 & \bf 0.9622 & 0.9407 & \bf 0.9564\\
\hline
\rule[-1ex]{0pt}{3.5ex} SRSIM DM1 & 0.9903  & 0.9827   & 0.9835   & 0.9946  & 0.9911  & 0.9869  & 0.9754   & 0.9713   & 0.9734   \\
\hline
\rule[-1ex]{0pt}{3.5ex} SRSIM DM2 & 0.9905  & 0.9895   & 0.9895   & 0.9944  & 0.9932  & 0.9900  & 0.9737   & 0.9762   & 0.9770   \\
\hline
\rule[-1ex]{0pt}{3.5ex} SRSIM DM3 & 0.9884  & 0.9829   & 0.9832   & 0.9933  & 0.9872  & 0.9838  & 0.9695   & 0.9681   & 0.9748   \\
\hline
\rule[-1ex]{0pt}{3.5ex} SRSIM DM4 & 0.9921 & \bf 0.9897 & 0.9896 & 0.9967 & \bf 0.9935 & 0.9921 & 0.9812 & \bf 0.9770 & 0.9795\\
\hline
\rule[-1ex]{0pt}{3.5ex} SRSIM DM5 & \bf 0.9932  & 0.9869   & \bf 0.9921  & \bf 0.9978  & 0.9893  & \bf 0.9984   & \bf 0.9842   & 0.9749   & \bf 0.9808  \\
\hline
\end{tabular}
\end{adjustbox}
\end{center}
\end{table}

\begin{table}[!tb]
\caption{ Comparison of efficiency of denoising by switching filter based on DetectionMethod(DM) 1-5 using quality measure (QM) PSNR, MSE, IFS, FSIM, SRSIM  with type of noise CI1-3, CT1-3, $p=0.2,0.3$.}
\label{tab:RuchTab3}
\begin{center}
\begin{adjustbox}{width=\textwidth}
\begin{tabular}{|l|c|c|c|c|c|c|c|c|c|}
\hline
\rule[-1ex]{0pt}{3.5ex} QM & CT1 0.2 & CT2 0.2 & CT3 0.2 & CI1 0.3 & CI2 0.3 & CI3 0.3 & CT1 0.3 & CT2 0.3 & CT3 0.3 \\
\hline
\rule[-1ex]{0pt}{3.5ex} PSNR DM1 & 23.572  & 25.547  & 25.126  & 18.246   & 19.246  & 20.556  & 18.393  & 20.720  & 19.674  \\
\hline
\rule[-1ex]{0pt}{3.5ex} PSNR DM2 & 23.026  & 24.637  & 25.683  & 19.206   & 19.245  & 20.316  & 19.426  & 21.052  & 22.297  \\
\hline
\rule[-1ex]{0pt}{3.5ex} PSNR DM3 & 26.004  & 25.753  & 26.528  & 16.929   & 18.738  & 20.153  & 21.077  & 22.704  & 23.182  \\
\hline
\rule[-1ex]{0pt}{3.5ex} PSNR DM4 & 26.436 & \bf 26.573 & 26.893 & 20.368 & \bf 19.533 & 21.808 & 21.404 & \bf 22.717 & 23.574\\
\hline
\rule[-1ex]{0pt}{3.5ex} PSNR DM5 & \bf 27.626 & 25.242 & \bf 28.549 & \bf 22.666 & 18.522 & \bf 23.262 & \bf 23.715 & 22.362 & \bf 25.01\\
\hline
\rule[-1ex]{0pt}{3.5ex} MSE DM1 & 285.64  & 181.28  & 199.74  & 973.69   & 773.49  & 572.01  & 941.25  & 550.90  & 700.85 \\
\hline
\rule[-1ex]{0pt}{3.5ex} MSE DM2 & 323.95  & 223.51  & 175.67  & 780.66   & 773.58  & 604.55  & 741.98  & 510.38  & 383.16 \\
\hline
\rule[-1ex]{0pt}{3.5ex} MSE DM3 & 163.16  & 172.87  & 144.62  & 1318.5   & 869.44  & 627.63  & 507.42  & 348.84  & 312.50 \\
\hline
\rule[-1ex]{0pt}{3.5ex} MSE DM4 & 155.95 & \bf 143.13 & 132.95 & 597.33 & \bf 723.92 & 428.82 & 502.49 & \bf 329.36 & 352.54 \\
\hline
\rule[-1ex]{0pt}{3.5ex} MSE DM5 & \bf 112.32 & 194.49 & \bf 90.82 & \bf 351.93 & 913.89 & \bf 306.84 & \bf 276.43 & 377.44 & \bf 205.15 \\
\hline
\rule[-1ex]{0pt}{3.5ex} IFS DM1 & 0.8963   & 0.8961   & 0.8931   & 0.7523    & 0.6922   & 0.7904   & 0.7687   & 0.8296   & 0.7655   \\
\hline
\rule[-1ex]{0pt}{3.5ex} IFS DM2 & 0.8887   & 0.8820   & 0.8981   & 0.7833    & 0.6799   & 0.8189   & 0.8099   & 0.8441   & 0.8371   \\
\hline
\rule[-1ex]{0pt}{3.5ex} IFS DM3 & 0.9358   & 0.9037   & 0.9122   & 0.7598    & 0.6741   & 0.7975   & 0.8519   & 0.8771   & 0.8577   \\
\hline
\rule[-1ex]{0pt}{3.5ex} IFS DM4 & 0.9367 & \bf 0.9163 & 0.9267 & 0.8186 & \bf 0.6972 & 0.8500 & 0.8539 & \bf 0.8783 & 0.8592\\
\hline
\rule[-1ex]{0pt}{3.5ex} IFS DM5 & \bf 0.9393  & 0.8968  & \bf 0.9341 & \bf 0.8824  & 0.6946  & \bf 0.884   & \bf 0.9029  & 0.8739  & \bf 0.8964\\
\hline
\rule[-1ex]{0pt}{3.5ex} FSIM DM1 & 0.9289 & 0.9349   & 0.9307   & 0.8373    & 0.8015   & 0.8663   & 0.8317   & 0.9084   & 0.8320   \\
\hline
\rule[-1ex]{0pt}{3.5ex} FSIM DM2 & 0.9277 & 0.9235   & 0.9398   & 0.8510    & 0.8006   & 0.8640   & 0.8599   & 0.9076   & 0.8929   \\
\hline
\rule[-1ex]{0pt}{3.5ex} FSIM DM3 & 0.9564 & 0.9436   & 0.9414   & 0.8106    & 0.7824   & 0.8562   & 0.8862   & 0.9247   & 0.9077   \\
\hline
\rule[-1ex]{0pt}{3.5ex} FSIM DM4 & 0.9566 & \bf 0.9482 & 0.9456 & 0.8778  & \bf 0.8211 & 0.8882 & 0.8867 & \bf 0.9254 & 0.9095\\
\hline
\rule[-1ex]{0pt}{3.5ex} FSIM DM5 & \bf 0.9584  & 0.9273  & \bf 0.9496 & \bf 0.9154  & 0.8194  & \bf 0.9052  & \bf 0.9265  & 0.9136  & \bf 0.9187\\
\hline
\rule[-1ex]{0pt}{3.5ex} SRSIM DM1 & 0.9752   & 0.9769   & 0.9701   & 0.9312    & 0.8973   & 0.9448   & 0.9315   & 0.9512   & 0.9289   \\
\hline
\rule[-1ex]{0pt}{3.5ex} SRSIM DM2 & 0.9751   & 0.9724   & 0.9735   & 0.9377    & 0.8957   & 0.9431   & 0.9446   & 0.9472   & 0.9570   \\
\hline
\rule[-1ex]{0pt}{3.5ex} SRSIM DM3 & 0.9817   & 0.9788   & 0.9725   & 0.9136    & 0.8752   & 0.9369   & 0.9557   & 0.9584   & 0.9623   \\
\hline
\rule[-1ex]{0pt}{3.5ex} SRSIM DM4 & 0.9830 & \bf 0.9826 & 0.9777 & 0.9517 & \bf 0.9125 & 0.9561 & 0.9563 & \bf 0.9621 & 0.9625\\
\hline
\rule[-1ex]{0pt}{3.5ex} SRSIM DM5 & \bf 0.9877  & 0.9708  & \bf 0.9782  & \bf 0.9651  & 0.9122  & \bf 0.9582  & \bf 0.9728  & 0.9512  & \bf 0.9688\\
\hline
\end{tabular}
\end{adjustbox}
\end{center}
\end{table}

\begin{figure}[!tb]
\begin{center}
\begin{tabular}{p{0.33\textwidth}p{0.33\textwidth}p{0.33\textwidth}}
\includegraphics[width=0.28\textwidth]{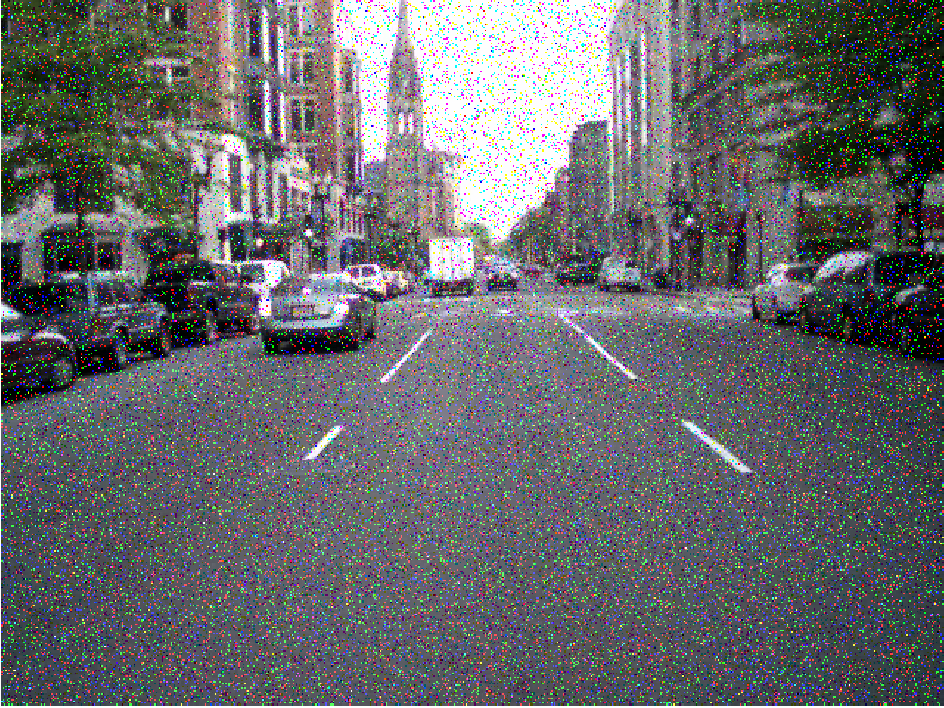} & \includegraphics[width=0.28\textwidth]{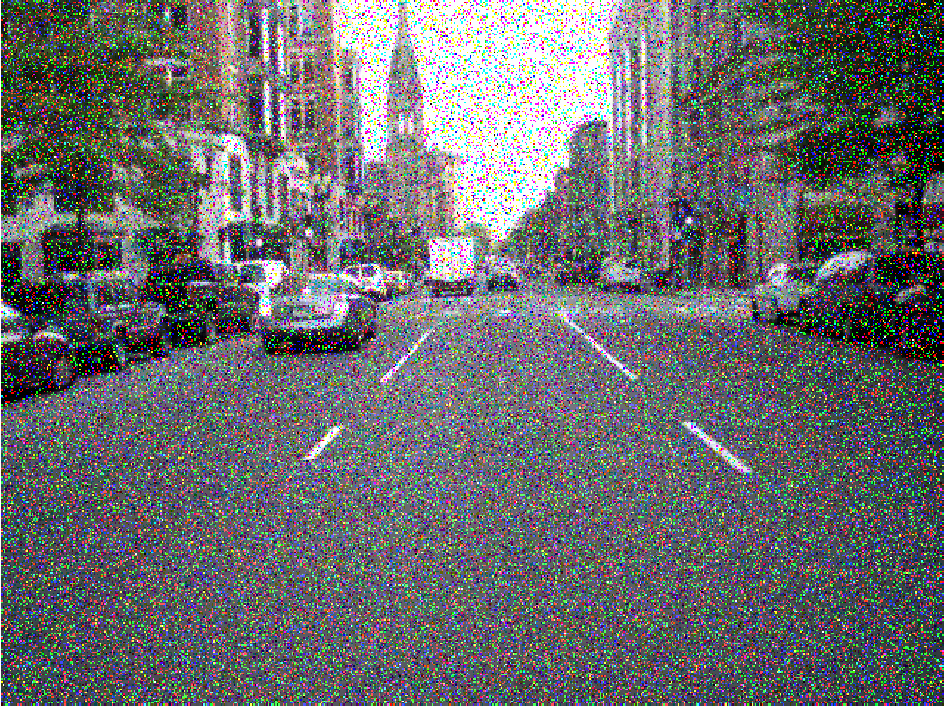} & \includegraphics[width=0.28\textwidth]{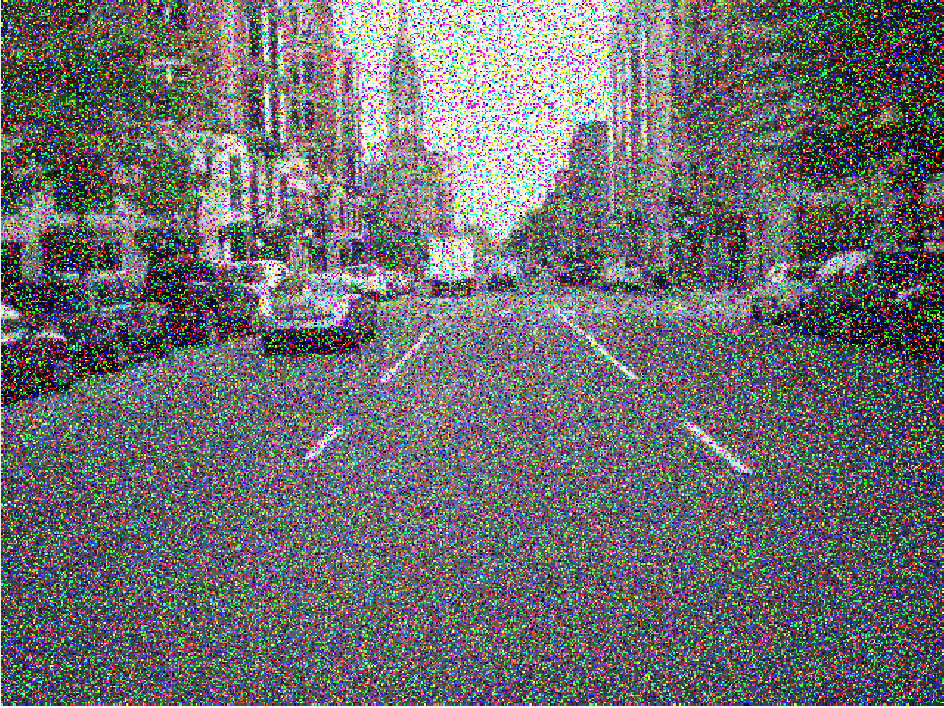}\\
\includegraphics[width=0.28\textwidth]{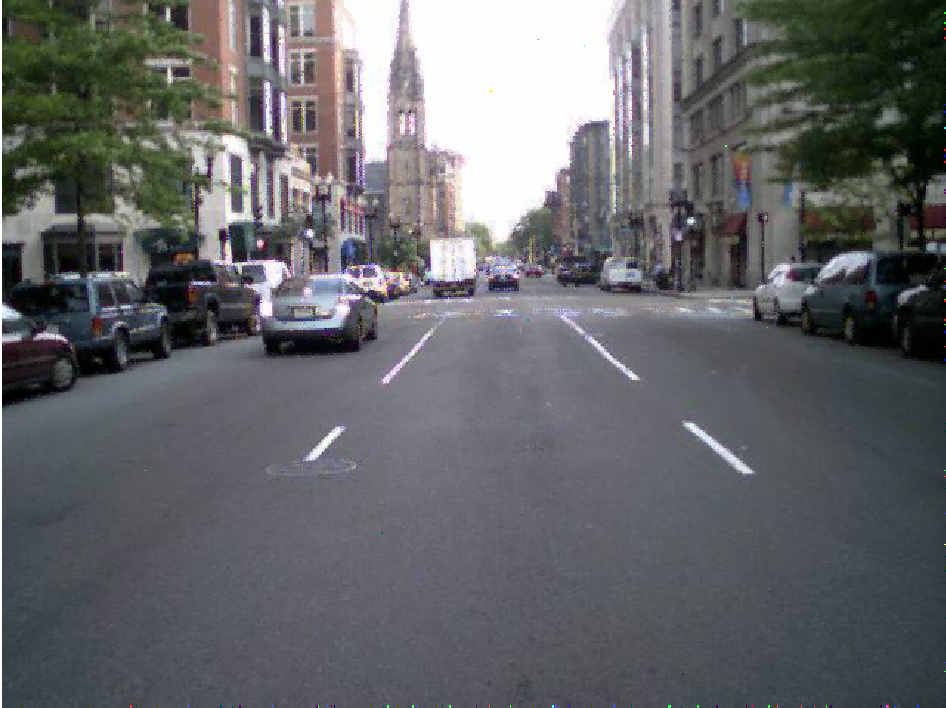} & \includegraphics[width=0.28\textwidth]{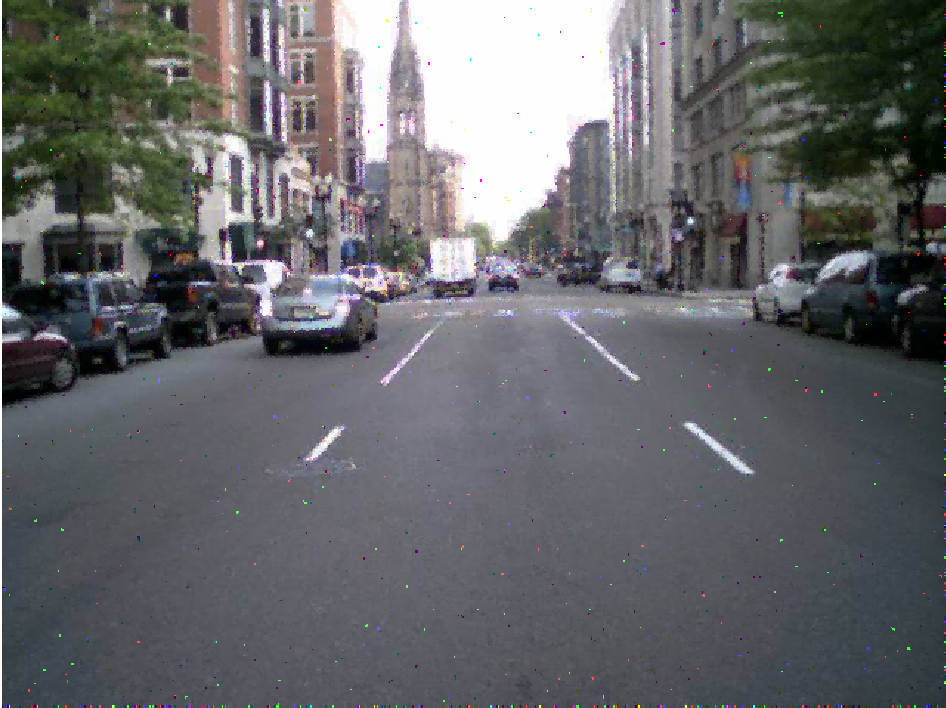} & \includegraphics[width=0.28\textwidth]{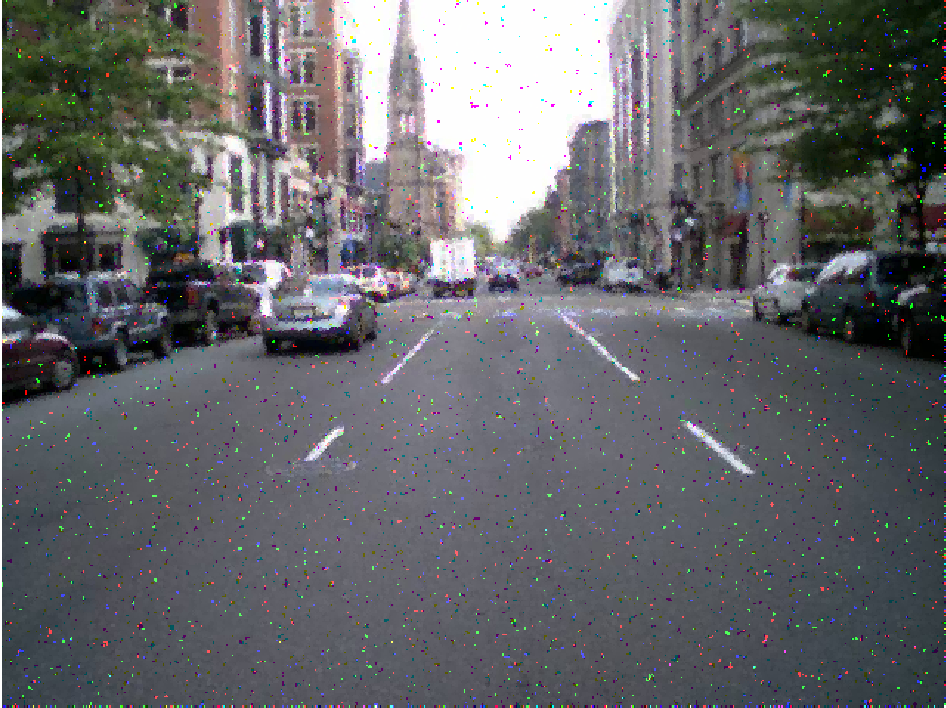}\\
\end{tabular}
\end{center}
\caption[example]
{ \label{fig:RuchFigToRes1} Results of denoising by switching filter based on DetectionMethod5 with type of noise CI1 $p=0.1,0.2,0.3$.}
\end{figure}

\begin{figure}[!tb]
\begin{center}
\begin{tabular}{p{0.33\textwidth}p{0.33\textwidth}p{0.33\textwidth}}
\includegraphics[width=0.28\textwidth]{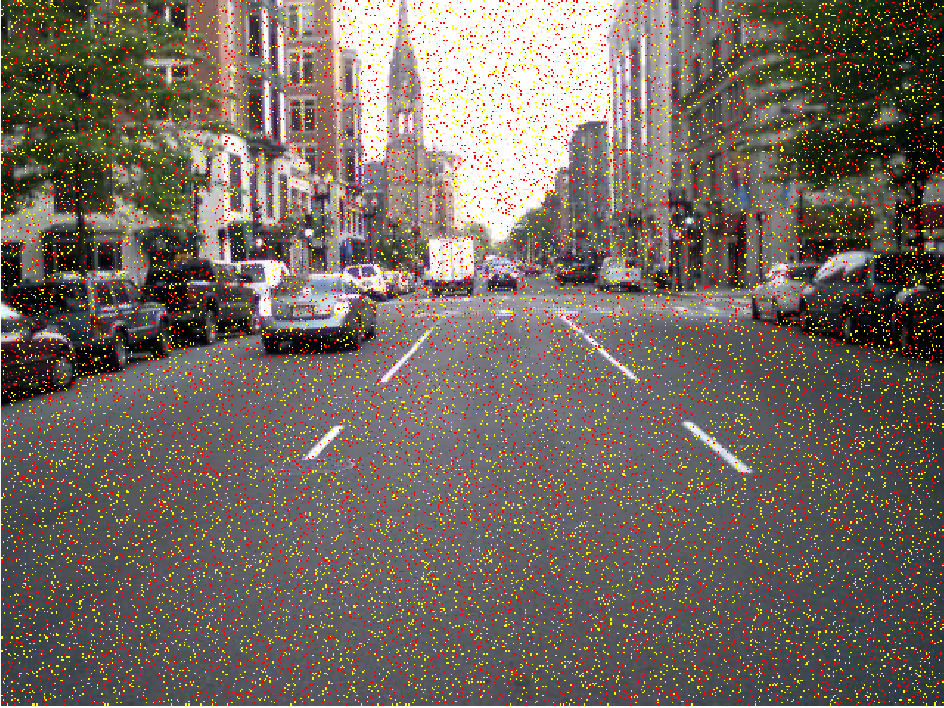} & \includegraphics[width=0.28\textwidth]{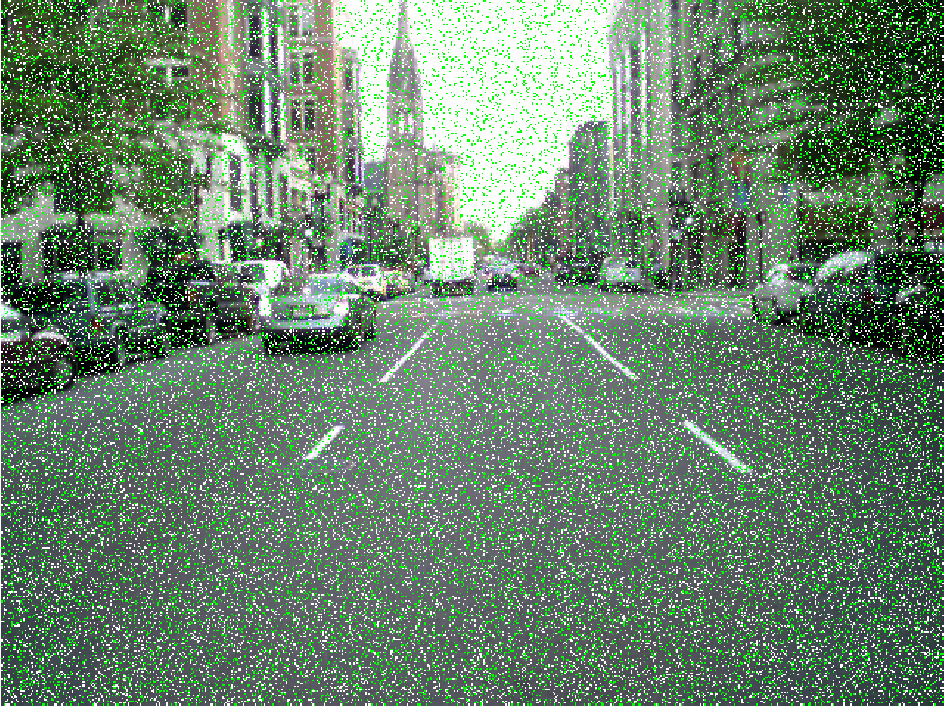} & \includegraphics[width=0.28\textwidth]{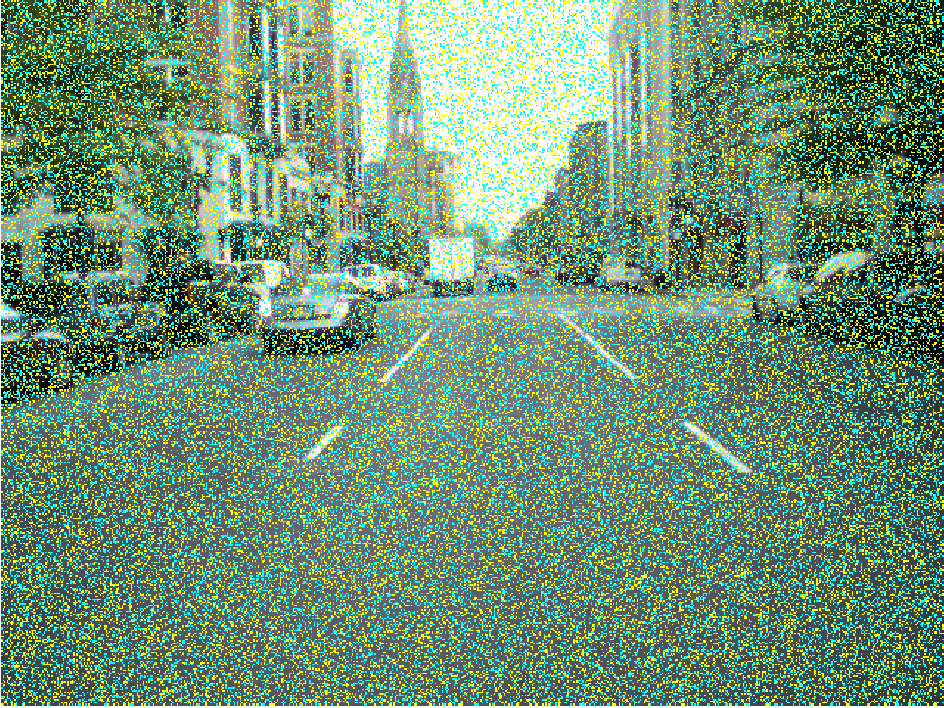}\\
\includegraphics[width=0.28\textwidth]{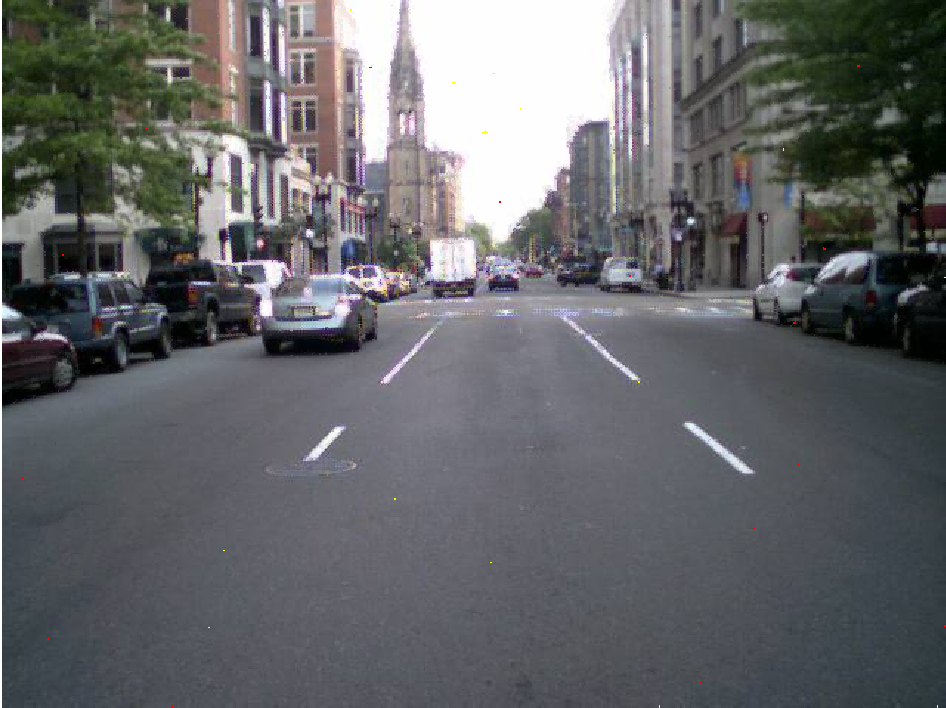} & \includegraphics[width=0.28\textwidth]{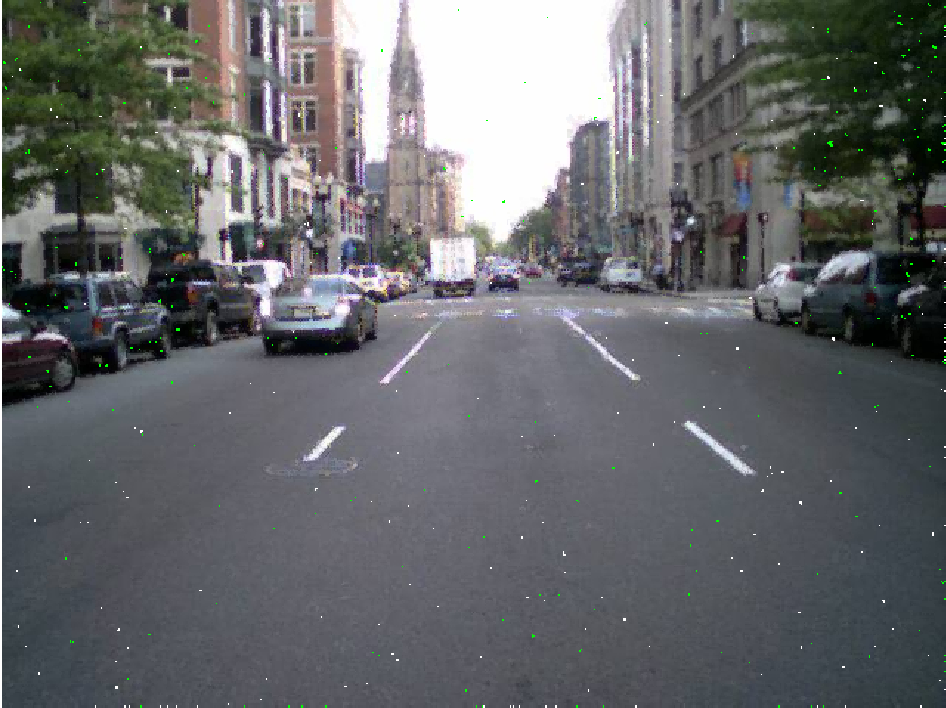} & \includegraphics[width=0.28\textwidth]{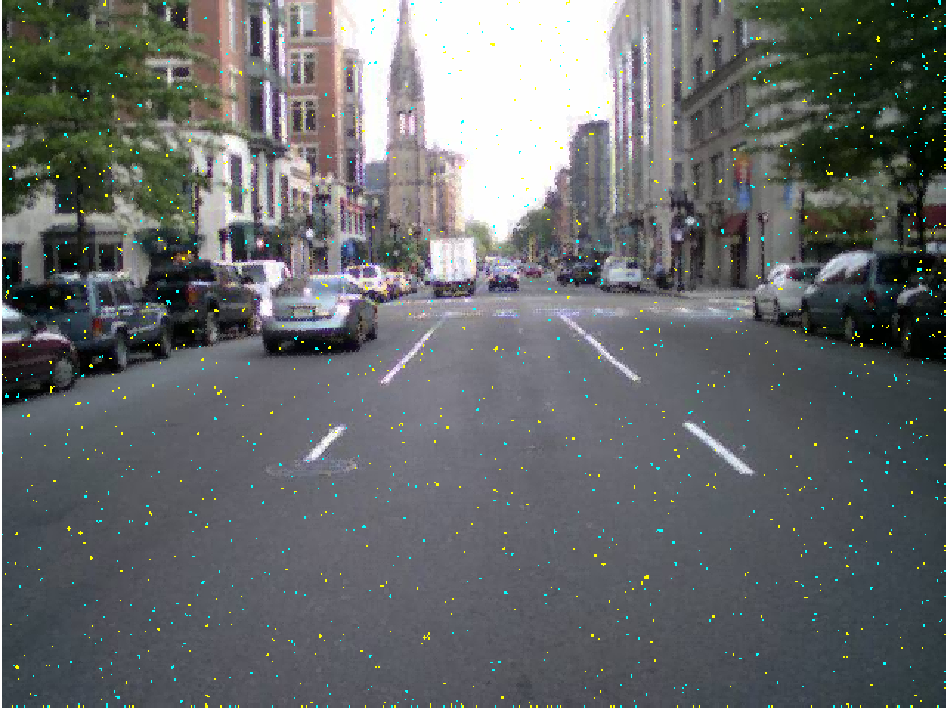}\\
\end{tabular}
\end{center}
\caption[example]
{ \label{fig:RuchFigToRes4} Results of denoising by switching filter based on DetectionMethod5 with type of noise CT1 $p=0.1,0.2,0.3$.}
\end{figure}

The result of denoising based on the proposed detection method \verb"DetectionMethod5" presented in Fig. \ref{fig:RuchFigToRes1}, and \ref{fig:RuchFigToRes4}.
We see that the proposed method with morphological filtering yields good results in terms of objective and subjective criteria.

Next we provide execution time of denoising algorithms with switching filter based on DetectionMethod5 with type of noise CI1-3, CT1-3, $p=0.1,0.2,0.3$. 20 experiments were carried out and the results are averaged. Table \ref{tab:RuchTabSpeed} show that the proposed algorithm with morphological filtering yields the best results in terms of execution time.

\begin{table}[!tb]
\caption{Execution time (seconds) of denoising algorithms based on DetectionMethod(DM) 1-5 with type of noise CI1-3, CT1-3, $p=0.1,0.2,0.3$.}
\label{tab:RuchTabSpeed}
\centering
\begin{tabular}{|l|c|c|c|c|c|}
\hline
\rule[-1ex]{0pt}{3.5ex} DM & DM1 & DM2 & DM3 & DM4 & DM5 \\
\hline
\rule[-1ex]{0pt}{3.5ex} Execution time & 16.12 & 16.02 & 7.43 & 7.23 & \textbf{0.06}\\
\hline
\end{tabular}
\end{table}

\hfill \break
\hfill \break
\hfill \break
%%%%%%%%%%%%%%%%%%%%%%%%%%%%%%%%%%%%%%%%%%%%%%%%%%%%%%%%%%%%%
\section{Conclusion}
In the paper, new noise detection techniques for switching filtering of impulse noise with morphological filtering were proposed. Computer simulation performed on test images contaminated by six noise models revealed a very high efficiency of the proposed method. The performance of the proposed algorithm was evaluated in terms of objective and subjective criteria of image restoration. With the help of computer simulation we showed that the proposed algorithm with morphological filtering can effectively remove impulse noise. Moreover the proposed algorithm is the faster among all tested algorithms.

%%%%%%%%%%%%%%%%%%%%%%%%%%%%%%%%%%%%%%%%%%%%%%%%%%%%%%%%%%%%%
\subsubsection*{Acknowledgements.}
The work was supported by the Ministry of Education and Science of Russian Federation, grant 2.1766.2014.

%%%%% References %%%%%
\bibliography{bibRuchay}
\bibliographystyle{splncs}

\end{document}